# Leveraging Large Language Models for Optimized Item Categorization using UNSPSC Taxonomy

Anmolika Singh and Yuhang Diao

Data Scientist, USA

## Abstract

*Effective item categorization is vital for businesses, enabling the transformation of unstructured datasets into organized categories that streamline inventory management. Despite its importance, item categorization remains highly subjective and lacks a uniform standard across industries and businesses. The United Nations Standard Products and Services Code (UNSPSC) provides a standardized system for cataloguing inventory, yet employing UNSPSC categorizations often demands significant manual effort. This paper investigates the deployment of Large Language Models (LLMs) to automate the classification of inventory data into UNSPSC codes based on Item Descriptions. We evaluate the accuracy and efficiency of LLMs in categorizing diverse datasets, exploring their language processing capabilities and their potential as a tool for standardizing inventory classification. Our findings reveal that LLMs can substantially diminish the manual labor involved in item categorization while maintaining high accuracy, offering a scalable solution for businesses striving to enhance their inventory management practices.*

## 1. Introduction

Businesses constantly seek to manage operations and policies for their Stock Keeping Units (SKUs), whether for pricing, prioritization, e-commerce, accounting, or resource distribution. As inventory grows, so do time investments and expenses [16]. Different item types require distinct strategies, enhancing the ease of processing and analysis. Therefore, there is a need for grouping products into categories.

Classifying items into product categories also becomes a fundamental aspect of inventory management, as it enhances the organization's efficiency of supply chain operations. This classification aids in inventory tracking, demand forecasting, and strategic decision-making. Moreover, in today's digitally evolved business landscape, the significance of standardized information transcends automated processes to encompass all electronically supported operations [3]. As a result, electronic product catalogues form the core of most inventory management practices, acting as hubs for regularly updated and effortlessly distributable product information, ensuring precision and uniformity throughout the supply chain [3]. Manufacturers, distributors, wholesalers, and resellers all gain substantial advantages from structured and categorized data to optimize processes and make well-informed decisions in business.

The implementation of common coding standards like UNSPSC plays a pivotal role in metamorphosing raw data into valuable, cohesive information [4]. By categorizing products into standardized classifications, UNSPSC enriches data utility for expenditure analysis, electronic commerce, logistics, and supplier sourcing [4]. This methodical classification supports efficient inventory management by empowering organizations to effectively oversee inventory levels and uncover opportunities for cost savings [4].

In this paper, we study the possible automation of this item categorization using a Large Language model such as OpenAI's GPT-4 [12]. Given the complexity inherent in categorization processes, employing LLMs through prompt engineering offers a promising avenue to streamline and enhance the accuracy of this task. LLMs mark a significant advancement in artificial intelligence and natural language processing, showcasing remarkable capabilities in comprehending and generating desired outcomes based on input.

## 2. LITERATURE REVIEW

The field of item categorization has evolved significantly, leveraging advancements in machine learning and natural language processing (NLP) to address the challenges of managing large and diverse inventories. This section reviews key contributions to the literature, including traditional approaches to inventory classification, the structure and application of the UNSPSC taxonomy, the transformative potential of large language models (LLMs) for automated categorization, and the emerging importance of prompt engineering in refining model outputs. By synthesizing prior research, this review highlights both the progress and ongoing challenges in the domain of item categorization.

### 2.1. Item Inventory Classification

Various approaches have been explored to automate the item classification problem. Shankar and Lin [15] have worked on applying supervised learning with known product categories and multi-class features to increase accuracy in the classification of products. However, this method worked on more broad generalized product categories. The results of Gottipati and Vauhkonen [5] experiments indicated that while a Chi-square model of feature selection worked with a small feature size set, for large feature set size Naïve Bayes gave the most accuracy with plain frequency-based Unigram model for feature extraction and LDA fared at an average level. In [19] the authors suggest Attention CNN (ACNN) model, for large-scale categorization of product titles into 35 top level categories.

### 2.2. UNSPSC Codes

The UNSPSC taxonomy is constructed as a tree structure with four levels called Segment, Family, Class and Commodity [17] Every level has a textual description as well as unique two digits. From the top of the tree to bottom, the category of each layer becomes one step more granular [8]. Combining all of these codes gives the unique UNSPSC code for the product.

1. Commodity: 12345678
2. Class: 12345600
3. Family: 12340000
4. Segment: 12000000

There have also been studies that focused particularly on classification using the UNSPSC code standard. In [1] the authors employed machine learning algorithms to develop the classification model with SVM having the best accuracy. In another study the researchers suggested using special clustering considering individual factors such as input or output of service operations to categorize services using UNSPSC Codes [9]. Karlsson and Karlstedt [8] in the thesis suggest that the optimal learning method for training is Support Vector Machines using inverse frequency class balancing, extracting features from the brand and title of products to give them an accurate UNSPSC taxonomy. However, the results of the study were inconclusive. Wolin [18] classifies products in UNSPSC-taxonomy by creating vectors based on words in each product category of the training set and computed the cosine similarity measure between an input product feature vector and all candidate categories.

### 2.3. Large Language Models

Recent advancements in Large Language Models offer a promising avenue for automated item categorization to be carried out efficiently and accurately. LLMs have become a powerful tool to apply Natural Language Processing (NLP) tasks and can be adapted for specific use case scenarios. In [6] the authors compare GPT-3 and their own model WHAM in the task of classifying job postings by the option of remote work at least one day per week.
In a research closer to our objectives of product classification [10] the GPT-3.5 model showed a high accuracy rate in categorizing products using Harmonized System (HS) nomenclature into duty tariff lines whereas traditional machine learning models performed well only with their training dataset.

### 2.4. Prompt Engineering

Prompt engineering is a critical technique for effectively utilizing large language models (LLMs). By crafting precise and contextually rich prompts, researchers can guide LLMs to produce accurate and relevant outputs. This process is essential for reducing ambiguity and enhancing the consistency of the model's responses. Brown et al. [2] demonstrated the efficiency of few-shot learning with carefully constructed prompts, highlighting that even a few examples can substantially improve model performance. Similarly, Raffel et al. [13] emphasized the importance of prompt format and structure in their exploration of transfer learning with the T5 model, showing that prompt variations can lead to different levels of success in text-to-text tasks.
In the context of item categorization, prompt engineering helps address challenges such as ambiguous item descriptions and the need for domain-specific knowledge. Schick and Schutze [14] explored the use of Cloze-style prompts, where the model fills in blanks within a sentence, for few-shot text classification. This technique demonstrated how prompts that mimic natural language questions elicit more accurate responses from LLMs. This approach is particularly relevant for UNSPSC categorization, where items must be matched to precise codes based on nuanced descriptions. By leveraging structured, contextual, and multi-turn prompts, researchers can enhance the accuracy of UNSPSC code assignment, thereby improving data standardization and procurement processes across various industries.

## 3. METHODOLOGY

This section outlines the approach used to explore the effectiveness of large language models (LLMs) for UNSPSC-based item categorization. The methodology involves leveraging a publicly available dataset, implementing structured prompts, and conducting experiments to evaluate the model's performance across various parameters.

## 3.1. Dataset

We use a public dataset of Purchase Orders from the State Government of California [11] From this source, we take a sample of 50,000 entries. The data provides comprehensive details about various purchase orders issued during this period. The dataset comprises 32 columns, capturing information about each purchase order, including:

- Item Details: Information about the items purchased, including item name, description, quantity, unit price, and total price.
- Supplier Details: Information about the supplier, including the supplier code, name, qualifications, and zip code.
- Classification Codes and UNSPSC: The classification codes and normalized UNSPSC associated with the items. Titles that further categorize the items based on their commodity, class, family, and segment.

    - 7658 Unique Commodity Codes
    - 1821 Unique Class Codes
    - 388 Unique Family Codes
    - 57 Unique Segment Codes

This dataset is rich in information and provides a valuable resource for item UNSPSC categorization. The inclusion of UNSPSC codes and detailed item descriptions makes it particularly suitable for studying the effectiveness of item UNSPSC categorization using large language models (LLMs).

## 3.2. Experiment Setup

### 3.2.1. Overview

The primary goal of this experiment is to leverage large language models (LLMs) such as Open AI's GPT-4 to classify items based on their names and descriptions using the UNSPSC. We evaluate the GPT-4 model's performance by comparing its classifications against a test dataset from the California State Government's purchase order data.

### 3.2.2. Dataset Preparation

The dataset is pre-processed to remove any incomplete records and to normalize item descriptions through methods such as text cleaning, and standardization.

### 3.2.3. Model Selection

The Azure Open AI API, specifically the GPT-4 model, is chosen for its advanced natural language processing capabilities. The GPT-4 model has the highest reasoning benchmarks at the time we ran the experiment [12]. It is accessed using an API key, with requests made to the specified endpoint.

### 3.2.4. Experimental Procedure

The experimental procedure involves a series of steps designed to leverage the Azure Open AI API for UNSPSC item categorization. Initially, input construction is performed for each item in the dataset, where a prompt is crafted to include both the item name and its description. These prompts are carefully formatted to ensure that the model receives all the necessary information for accurate classification. Following input construction, POST requests are sent to the API using these prompts. Each request is structured to query the model and retrieve a prediction for the UNSPSC code corresponding to the item. Once the responses are received, they are parsed to extract the predicted UNSPSC codes. This parsing process ensures that the outputs are correctly interpreted and stored for subsequent evaluation against the actual UNSPSC codes in the dataset. This systematic approach allows for a thorough assessment of the model's performance in classifying items based on their names and descriptions.

### 3.2.5. Prompt Building

For the study, we build a generic prompt that guides the LLM to assign UNSPSC Codes from the product name and item descriptions.

> **Prompt 1 - Generic Prompt**
> You will receive a product name and description. Your task is to classify the product into the appropriate UNSPSC category. Provide your output as the UNSPSC code only.

Next, we implemented a Cloze-style prompt, a widely used technique in natural language processing, where part of the input is left blank for the model to fill. This style is effective for classification tasks because it frames the problem as a question with a single, context-dependent answer. In our case, we structured the prompt to mimic a natural query for the UNSPSC code, expecting it to improve accuracy by reducing ambiguity in the model's output.

> **Prompt 2 - Cloze-style Prompt**
> The appropriate UNSPSC code (a numerical code) for a product named
> '{item_name}' described as '{item_description}' is:"

In another approach, we used a long prompt-engineered statement that provides context to the LLM by including example usage. The following are examples of product name and description included in the user prompt to guide the model:

> **Prompt 3 - Engineered Prompt**
> You will receive a product name and description. Your task is to classify the product into the appropriate UNSPSC category. Provide your output as the UNSPSC code only.
>
> **Example 1:**
> *User Prompt:*
> *Product Name: HP LaserJet Pro M404dn*
> *Description: Laser printer, black and white, 40 pages per minute.*
> *Expected Output: 43212110*
>
> **Example 2:**
> *User Prompt:*
> *Product Name: Dell Latitude 7420*
> *Description: Business laptop with 14-inch screen and Intel i7 processor.*
> *Expected Output: 43211503*
>
> **Example 3:**
> *User Prompt:*
> *Product Name: 3M Scotch Magic Tape*
> *Description: Invisible tape for office use, 1 inch by 1000-inch roll.*
> *Expected Output: 31201512*

With these three prompts, we run our experiments to test the impact of prompt creation on the accuracy of the LLM.

### 3.3. Result Metrics

In the experiment, we employ a generic prompt, a Cloze-style prompt, and an engineered prompt to test the model's performance. For each prompt, the model is evaluated at 3 different levels of temperatures. The temperature parameter of the LLMs controls the randomness of its output. A higher temperature allows for more diverse and creative responses, as the model is encouraged to explore a broader range of possibilities. However, this increased diversity may also lead to less accurate or inconsistent predictions. Conversely, lower temperature values reduce randomness, resulting in more focused and deterministic predictions. In our case, because we are performing a classification task, we expect that lower temperatures will yield higher accuracy, as there is less variation in the output, leading to more deterministic and reliable predictions.

The evaluation of the model's performance is based on accuracy, which measures the proportion of correct product categorization made by the model. The accuracy is calculated for each hierarchy layer of the UNSPSC code assigned by the LLM that is the commodity, the Class, the Family, and the Segment. In this experiment, the model's performance is assessed across the following criteria:

1. Prompt
2. Temperature
3. UNSPSC Hierarchy

International Journal on Cybernetics & Informatics (IJCI) Vol.13, No.6, December2024

## 4. RESULT ANALYSIS

The experiment evaluated the prediction accuracy of LLMs, to classify items into UNSPSC codes at different levels of the hierarchy, ranging from commodity to segment level. The results, as presented in Table 1, Table 2, and Table 3, provide insights into the performance of different prompts and temperature settings.
Across all tested prompts, it was observed that the prediction accuracy generally increased as we moved up the hierarchy of UNSPSC code from commodity to segment level.

Amongst the tested prompts, Prompt 3 consistently performed the best in terms of accuracy. Prompt 3 was particularly effective in handling cases where the model encountered inadequate information, as it provided a statement informing about the insufficiency rather than supplying a potentially incorrect result.

A temperature of '0' consistently yielded the best accuracy across all the results. This suggests that a lower temperature parameter results in more deterministic and accurate predictions.

Table 1. Prompt 1 Matrix

| Temperature | Accuracy Commodity | Accuracy Class | Accuracy Family | Accuracy Segment |
|---|---|---|---|---|
| 1 | 9.73% | 24.88% | 34.72% | 49.09% |
| 0.50 | 10.09% | 25.64% | 36.08 % | 49.59% |
| 0 | 10.20% | 25.88% | 35.75% | 49.88% |

Table 2. Prompt 2 Matrix

| Temperature | Accuracy Commodity | Accuracy Class | Accuracy Family | Accuracy Segment |
|---|---|---|---|---|
| 1 | 11.07% | 27.44% | 38.80% | 53.20% |
| 0.50 | 11.29% | 28.01% | 39.27 % | 53.63% |
| 0 | 11.45% | 28.30% | 39.52% | 53.73% |

Table 3. Prompt 3 Matrix

| Temperature | Accuracy Commodity | Accuracy Class | Accuracy Family | Accuracy Segment |
|---|---|---|---|---|
| 1 | 10.34% | 27.74% | 39.00% | 53.10% |
| 0.50 | 10.73% | 28.77% | 40.23 % | 54.49% |
| 0 | 10.80% | 29.01% | 40.31% | 54.59% |

Overall, the experiment results highlight the importance of prompt design and temperature setting in achieving accurate UNSPSC code predictions. The findings suggest that utilizing Prompt 3 with a temperature of '0' can lead to improved accuracy, particularly at higher levels of the UNSPSC hierarchy. These insights can be valuable for researchers and practitioners working on enhancing the prediction accuracy of UNSPSC codes.

## 5. CONCLUSION

In this research, we conducted an extensive experiment to evaluate the efficiency of LLM's capability to categorize items into UNSPSC codes. Our findings indicate that the prediction accuracy improves as we move up the hierarchy from the commodity to the segment level.

Prompt 3 consistently outperformed the other tested prompts, particularly when the temperature was set to '0'. This configuration not only yielded the highest accuracy but also effectively handled cases with inadequate information. This indicates that the prompt design plays a crucial role in obtaining accurate predictions.

When comparing our results to earlier work on the classification of products and services using UNSPSC codes [7], several key observations emerge. The baseline method in the earlier work utilized Naive Bayes probabilistic models and achieved promising results at the class level. Subsequent improvements using word2vec with averaging and logistic regression, as well as FastText and RoBERTa, demonstrated enhanced performance, with RoBERTa achieving the highest mF1 and wF1 scores of 0.912 and 0.904, respectively. However, the authors reported a significant drop in performance to mF1 scores of 50%-55% when tested on a larger, more diverse dataset.

Our experiment, which focused on a larger and immensely diverse dataset, showed that the accuracy at the class level reached up to 40.31%. Additionally, our results at the segment level, with accuracy reaching up to 54.59%, indicate strong performance at higher levels of the UNSPSC hierarchy.

In conclusion, our research contributes valuable insights into the effectiveness of LLMs in the prediction of UNSPSC codes, emphasizing the significance of prompt design and temperature settings. Our results are encouraging, particularly at higher levels of the UNSPSC hierarchy. LLMs have proven to be a highly effective resource for item categorization, significantly reducing time and manual effort. Future research could explore the impact of prompt engineering and temperature adjustments in other classification systems beyond UNSPSC, as well as further optimizing LLMs for more granular categorization tasks.

## 6. FURTHER WORK

The rapid evolution of Large Language Models (LLMs) opens new avenues for advancing item categorization tasks. Our experiment utilized Open AI's GPT-4 recognized as the best-performing model available at the time of this study, ensuring our findings were benchmarked against the most capable technology of the period. However, with the continuous release of more advanced models and other emerging architectures, it is crucial to revisit these experiments with newer LLMs. These upcoming models are expected to offer enhanced contextual understanding, improved reasoning capabilities, and greater adaptability to diverse datasets, potentially surpassing the results achieved in this study.

While this research focused on UNSPSC codes, its methodologies are transferable to other classification systems. Testing newer LLMs on frameworks like the North American Industry Classification System (NAICS) or Harmonized System (HS) codes could demonstrate their adaptability and broader application. Moreover, hybrid approaches that integrate traditional machine learning models with LLMs could leverage the strengths of both paradigms to further improve classification efficiency and accuracy.

## ACKNOWLEDGEMENTS

We would like to express our gratitude to everyone who provided support and guidance throughout the development of this research. The authors, Anmolika Singh and Yuhan Diao, thank each other for their collaboration and dedication in bringing this research to fruition.

# REFERENCES

[1] Bello Abdullahi, Yahaya Makarfi Ibrahim, Ahmed Doko Ibrahim, Kabir Bala, Yusuf Ibrahim, and Muhammad Aliyu Yamusa. Development of machine learning models for categorisation of nigerian government's procurement spending to unspsc procurement taxonomy. International Journal of Procurement Management, 19(1):106–121, 2024.

[2] Tom Brown, Benjamin Mann, Nick Ryder, Melanie Subbiah, Jared D Kaplan, Prafulla Dhariwal, Arvind Neelakantan, Pranav Shyam, Girish Sastry, Amanda Askell, et al. Language models are few-shot learners. Advances in neural information processing systems, 33:1877–1901, 2020.

[3] Frank Brüggemann and Ursula Hübner. From Product Identification to Catalog Standards, pages 127–154. Springer London, London, 2008.

[4] A.M. Fairchild and B. de Vuyst. Coding standards benefiting product and service information in e-commerce. In Proceedings of the 35th Annual Hawaii International Conference on System Sciences, pages 3201–3208, 2002.

[5] Srinivasu Gottipati. E-commerce product categorization srinivasu gottipati and mumtaz vauhkonen, 2012.

[6] Stephen Hansen, Peter John Lambert, Nicholas Bloom, Steven J Davis, Raffaella Sadun, and Bledi Taska. Remote work across jobs, companies, and space. Technical report, National Bureau of Economic Research, 2023.

[7] Ihor Hrysha and Samuel Grondahl. Large-scale product classification to recommend suppliers in procurement systems.

[8] Mikael Karlsson and Anton Karlstedt. Product classification-a hierarchical approach. LU-CS-EX 2016-31, 2016.

[9] Qianhui Liang, Peipei Li, Patrick CK Hung, and Xindong Wu. Clustering web services for automatic categorization. In 2009 IEEE International Conference on Services Computing, pages 380–387. IEEE, 2009.

[10] Ignacio Marra de Artiñano, Franco Riottini Depetris, and Christian Volpe Martincus. Automatic product classification in international trade: Machine learning and large language models. Technical report, IDB Working Paper Series, 2023.

[11] State of California. Purchase order data, 2024.

[12] OpenAI. gpt-4-0613. https://platform.openai.com/docs/models/gpt-4-turbo-and-gpt-4, 2023. Accessed: June 11, 2024.

[13] Colin Raffel, Noam M. Shazeer, Adam Roberts, Katherine Lee, Sharan Narang, Michael Matena, Yanqi Zhou, Wei Li, and Peter J. Liu. Exploring the limits of transfer learning with a unified text-to-text transformer. J. Mach. Learn. Res., 21:140:1–140:67, 2019.

[14] Timo Schick and Hinrich Schütze. Exploiting cloze questions for few shot text classification and natural language inference. arXiv preprint arXiv:2001.07676, 2020.

[15] Sushant Shankar and Irving Lin. Applying machine learning to product categorization. Department of Computer Science, Stanford University, 2011.

[16] Banu Soylu and Bahar Akyol. Multi-criteria inventory classification with reference items. Computers & Industrial Engineering, 69:12–20, 2014.

[17] unspsc org. United nations standard products and services code® (un-spsc®), 2022.

[18] Ben Wolin. Automatic classification in product catalogs. In Proceedings of the 25th annual international acm sigir conference on research and development in information retrieval, pages 351–352, 2002.

[19] Yandi Xia, Aaron Levine, Pradipto Das, Giuseppe Di Fabbrizio, Keiji Shinzato, and Ankur Datta. Large-scale categorization of japanese product titles using neural attention models. In Proceedings of the 15th Conference of the European Chapter of the Association for Computational Linguistics: Volume 2, Short Papers, pages 663–668, 2017.

**AUTHORS**

**Anmolika Singh** received the B.S. degree in Applied Data Sciences from the Pennsylvania State University, University Park, in 2021.

From 2021 to 2024, she was part of the prestigious Stanley Leadership Program, where she served as an Artificial Intelligence and Data Analytics Associate working on multiple high impact projects. She currently serves as a Data Scientist at Stanley Black & Decker's Industrial Business Unit, where she extracts valuable insights from industrial data to inform strategic decision-making and process improvements.

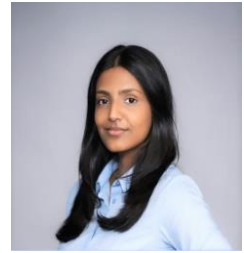

She is the author of several research articles, and her interests include applying data science techniques to solve real-world problems.

**Yuhang Diao** earned a B.S. degree in Econometrics from Ohio State University in 2019, followed by a master's degree in data Analytics Engineering from Northeastern University in 2021.

From 2022 to 2024, he participated in the Stanley Black & Decker Leadership Program as a Data Analyst. During this time, he worked on developing predictive models and optimizing data-driven decision-making processes. Currently, he is a Data Scientist in the Industrial Segment, where he has implemented advanced machine learning techniques, such as classification algorithms and neural networks, to drive innovation and operational efficiency. His work has contributed significantly to the development of smart storage solutions, including automated inventory management systems that leverage AI to enhance accuracy and performance.

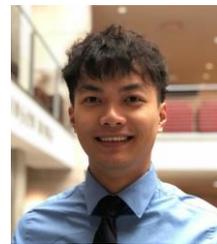